\documentclass[11pt]{article}

\usepackage[letterpaper,
            top=1.1in, bottom=1.1in,
            left=1.15in, right=1.15in]{geometry}

\usepackage{microtype}          
\usepackage{lmodern}            
\usepackage[T1]{fontenc}
\usepackage[utf8]{inputenc}

\usepackage{amsmath}
\usepackage{amsfonts}
\usepackage{amssymb}

\usepackage{graphicx}
\usepackage{booktabs}
\usepackage{multirow}
\usepackage{array}
\usepackage{tabularx}
\usepackage{makecell}

\usepackage[font=small, labelfont=bf, labelsep=period]{caption}

\usepackage{titlesec}
\titleformat{\section}{\large\bfseries}{\thesection.}{0.5em}{}
\titleformat{\subsection}{\normalsize\bfseries}{\thesubsection.}{0.5em}{}
\titleformat{\subsubsection}{\normalsize\itshape}{\thesubsubsection.}{0.5em}{}
\titlespacing*{\section}{0pt}{1.4ex plus .4ex minus .2ex}{0.6ex plus .1ex}
\titlespacing*{\subsection}{0pt}{1.2ex plus .3ex minus .2ex}{0.4ex plus .1ex}
\titlespacing*{\subsubsection}{0pt}{1.0ex plus .2ex minus .2ex}{0.3ex plus .1ex}

\usepackage{fancyhdr}
\fancypagestyle{firstpage}{%
  \fancyhf{}
  
  \fancyfoot[C]{\small\thepage}
}

\usepackage[colorlinks=true,
            linkcolor=black,
            citecolor=blue,
            urlcolor=blue]{hyperref}
\usepackage{url}

\usepackage[style=numeric,
            sorting=none,
            backend=biber,
            maxnames=15,
            giveninits]{biblatex}

\usepackage{abstract}

\usepackage{xspace}
\usepackage{indentfirst}  
\usepackage{flafter}  

\newcommand{\supref}[1]{\textsuperscript{\normalfont#1}}

\begin{document}
\thispagestyle{firstpage}

\begin{center}
  {\Large\bfseries
    Domain-Adapted Fine-Tuning of ECG Foundation Models\\[3pt]
    for Multi-Label Structural Heart Disease Screening\par}

  \vspace{0.8em}

  {\small
    Duc N. Do\supref{1,2}, Minh N. Do\supref{2}, Dang Nguyen\supref{3}, Khanh T.Q. Le\supref{1}, Khoa D. Pham\supref{1}, Hung N. Huynh\supref{1},\\[2pt]
    Phi Pham-Van-Hoang\supref{1}, Quan K. Huynh\supref{1}, Ramez M. Odat\supref{4}, Perisa Ashar\supref{5}, Ethan Philip Lowder\supref{6}, \\[2pt]
    Minh H.N. Le\supref{1}, Hoang Le\supref{1}, Phat V.H. Nguyen\supref{1}, Quan Le\supref{1}, Jacques Kpodonu\supref{7}, and Phat K. Huynh\supref{1,*}
  \par}

  \vspace{0.6em}

  {\footnotesize
    \supref{1}PASSIO Laboratory, North Carolina A\&T State University, Greensboro, NC, USA\\[1pt]
    \supref{2}Department of Computer Science, University of Manitoba, Winnipeg, MB, Canada\\[1pt]
    \supref{3}Harvard T.H. Chan School of Public Health, Harvard University, Boston, MA, USA\\[1pt]
    \supref{4}Department of Medicine, Jordan University of Science and Technology, Irbid, Jordan\\[1pt]
    \supref{5}Department of Biomedical Engineering, Duke University, Durham, NC, USA\\[1pt]
    \supref{6}Harvard Medical School, Boston, MA, USA\\[1pt]
    \supref{7}Beth Israel Deaconess Medical Center, Harvard Medical School, Boston, MA, USA\\[4pt]
    \supref{*}Correspondence: \href{mailto:pkhuynh@ncat.edu}{\texttt{pkhuynh@ncat.edu}}
  \par}

  \vspace{0.5em}

  {\small\itshape
    Accepted at the 39th Canadian Conference on Artificial Intelligence (Canadian AI 2026). \\
    To appear in \textit{Proceedings of Machine Learning Research}, vol.~318.\par}
\end{center}

\vspace{0.5em}

\begin{abstract}
Transthoracic echocardiography is the reference standard for confirming structural heart disease (SHD), but its use as a first-line screening modality is limited by cost, workflow burden, and specialist availability. This study investigated whether open pretrained electrocardiogram (ECG) models can support echo-confirmed multi-label SHD detection using the public EchoNext Mini-Model benchmark. We focused on six moderate-or-greater echocardiography-derived abnormalities spanning reduced left ventricular ejection fraction, increased left ventricular wall thickness, aortic stenosis, mitral regurgitation, tricuspid regurgitation, and right ventricular systolic dysfunction. Under a common experimental pipeline, we compared engineered ECG features with gradient boosting, end-to-end waveform learning from scratch, and transfer from open ECG foundation models. We then evaluated continued in-domain self-supervised adaptation of ECG-FM on EchoNext waveforms followed by selective supervised fine-tuning, with emphasis on the trade-off between discrimination and adaptation cost. Among the evaluated configurations, the adapted ECG-FM models achieved the strongest overall performance. Across adaptation depths, peak macro-AUROC and macro-AUPRC reached 0.8509 and 0.4297, respectively, while a more parameter-efficient operating point preserved nearly identical AUROC (0.8501) and achieved the highest fixed-threshold macro-F1 (0.3691). Late fusion of the release-provided covariates did not improve threshold-independent discrimination, and the evaluated low-rank adaptation (LoRA) configuration, alternative foundation backbones, and mixture-of-foundation-model strategies did not surpass the best adapted single-backbone operating points. These findings indicate that, for ECG-based case finding and echocardiography triage, the most effective transfer strategy is to combine target-domain self-supervised adaptation with selective supervised updating of a pretrained ECG backbone.

\end{abstract}

\vspace{0.5em}

\section{Introduction}
\label{sec:introduction}

Structural heart disease (SHD) includes clinically important abnormalities of cardiac structure and function, such as ventricular systolic dysfunction, ventricular hypertrophy, and moderate-or-greater valvular disease. Transthoracic echocardiography (TTE) is the standard imaging test used to confirm and grade these abnormalities, but broad first-line use of echocardiography is limited by cost, equipment, workflow burden, and specialist availability. In practice, referral pathways therefore rely on lower-cost front-line tests to identify patients who may warrant confirmatory TTE. This makes ECG-based case-finding and echocardiography triage a natural setting for machine learning methods that can prioritize imaging without adding major clinical burden \cite{siontis2021aiecg,ulloa2022rechommend,poterucha2025echonext}.

The 12-lead electrocardiogram (ECG) is especially attractive for this purpose because it is inexpensive, ubiquitous, and acquired in a standardized format. Deep learning studies have shown that raw ECG waveforms contain predictive information beyond conventional rule-based interpretation, enabling end-to-end learning for arrhythmia and diagnostic classification \cite{hannun2019arrhythmia,ribeiro2020automatic}. Public ECG benchmarks have further improved reproducibility and made it possible to compare model architectures and training strategies under common protocols \cite{wagner2020ptbxl,strodthoff2021ptbxl}. However, many widely used ECG benchmarks are built around rhythm, conduction, or report-derived ECG interpretation labels as opposed to structural phenotypes confirmed by echocardiography, which limits their direct relevance for echocardiography triage.

Prior work has established the feasibility of detecting echo-confirmed structural abnormalities from ECGs. Attia \emph{et al.} showed that AI-enabled ECG analysis can identify reduced left ventricular ejection fraction (LVEF) \cite{attia2019lvef}. Elias \emph{et al.} extended this idea to moderate-or-severe valvular disease \cite{elias2022valvular}. Ulloa-Cerna \emph{et al.} proposed rECHOmmend as a composite ECG-based model to prioritize echocardiography for patients at elevated risk of clinically significant, previously unrecognized SHD \cite{ulloa2022rechommend}. More recently, the EchoNext study reported large-scale SHD detection from ECGs and released the EchoNext Mini-Model PhysioNet dataset, enabling public evaluation on paired ECG and echocardiographic data \cite{poterucha2025echonext,elias2025echonext,goldberger2000physionet}.

What remains unclear is not whether ECGs contain signal relevant to SHD, but how open pretrained ECG encoders should be adapted for echo-confirmed, multi-label SHD prediction. Existing studies have mainly emphasized task-specific supervised models, whereas the rapid emergence of self-supervised and foundation ECG models raises a more practical transfer question. On a shared public SHD benchmark, \emph{when is a frozen encoder with a lightweight classifier sufficient, when is full fine-tuning justified, and when does an intermediate strategy provide a better balance between discrimination and adaptation cost?}

Self-supervised learning (SSL) offers a natural starting point because it learns transferable waveform representations from large unlabeled corpora. In ECG analysis, SSL approaches such as contrastive learning of cardiac signals (CLOCS) have improved downstream generalization across tasks \cite{kiyasseh2021clocs}. Open ECG foundation models (FMs), including ECG-FM \cite{mckeen2025ecgfm}, ECGFounder \cite{li2025ecgfounder}, and HuBERT-ECG \cite{coppola2024hubertecg}, have made this transfer setting testable. Yet performance in SHD detection may depend on two factors that are rarely disentangled: (i) mismatch between the pretraining distribution and the target clinical population, and (ii) the extent to which pretrained backbone weights are updated during supervised learning.

This study presents a controlled empirical evaluation of how open pretrained ECG models should be adapted for echo-confirmed multi-label SHD detection. Using EchoNext Mini-Model, we focus on six moderate-or-greater structural abnormalities spanning ventricular dysfunction, hypertrophy, and clinically significant valvular disease, and compare engineered ECG features with gradient boosting, end-to-end waveform learning from scratch, and transfer from open ECG FMs under a common experimental pipeline. We further isolate the contributions of continued in-domain self-supervised adaptation and layer-wise selective fine-tuning, asking whether these transfer choices yield a better discrimination--efficiency trade-off than frozen transfer or broader end-to-end adaptation. The paper contributes a rigorous practical assessment of which transfer decisions are most effective for clinically relevant multi-label SHD detection from routine 12-lead ECGs.

\section{Methods}
\label{sec:materials_methods}

\begin{figure}[!htbp]
    \centering
    \includegraphics[width=\linewidth]{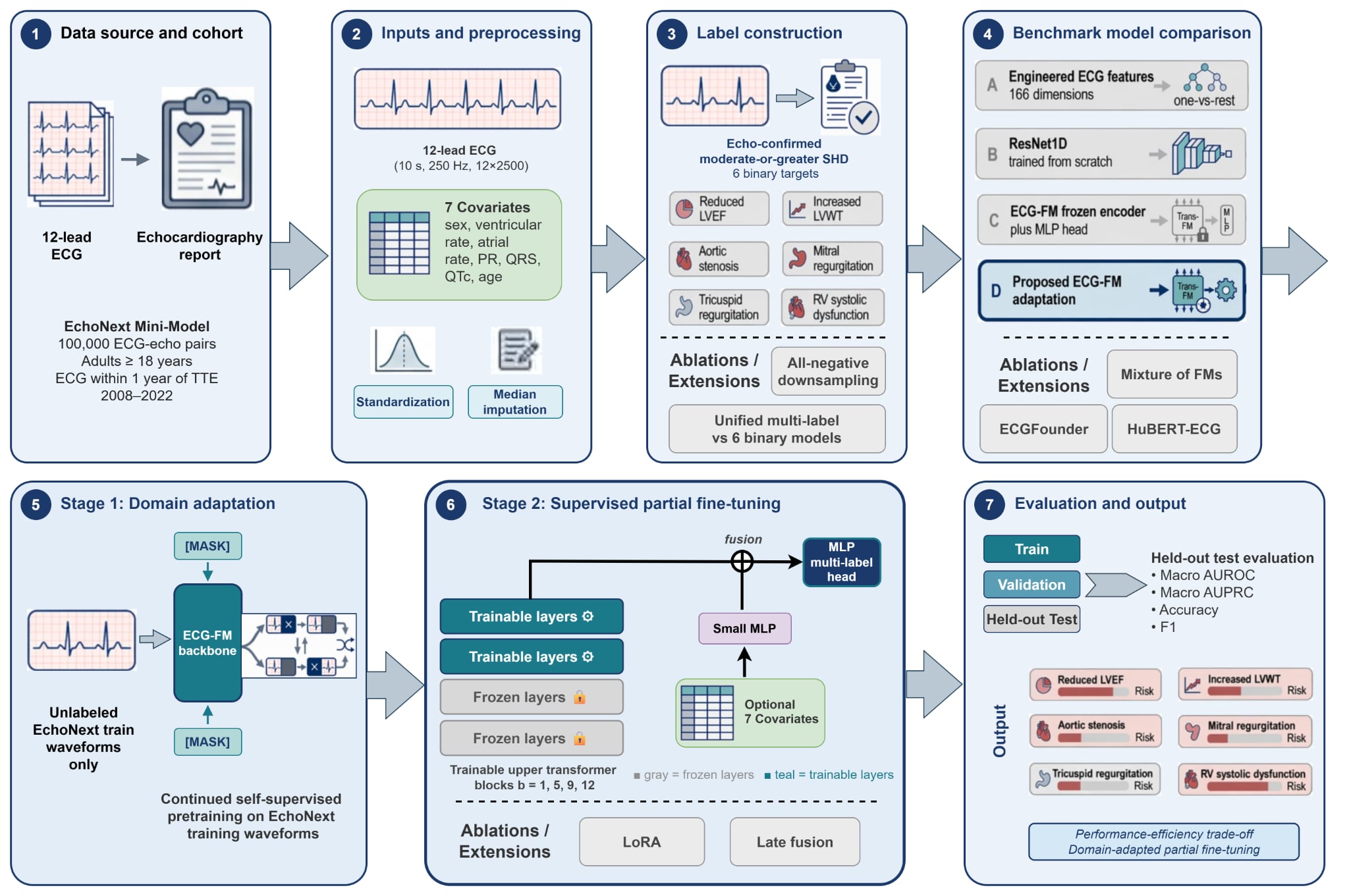}
    \caption{Overall methodological workflow for multi-label SHD detection from 12-lead ECGs. The framework begins with the EchoNext Mini-Model cohort and released waveform and covariate inputs, proceeds through construction of six moderate-or-greater echocardiography-confirmed endpoints, and compares engineered-feature, from-scratch waveform, and pretrained FM baselines. The primary transfer setting applies continued in-domain self-supervised adaptation of ECG-FM on EchoNext waveforms followed by supervised fine-tuning at varying adaptation depths. The workflow also includes late-fusion, LoRA, alternative-backbone, mixture-of-foundation-model, binary-decomposition, and all-negative downsampling ablations.}
    \label{fig:method_overview}
\end{figure}

The overall study workflow is illustrated in Figure~\ref{fig:method_overview}. 

\subsection{Data source and cohort}
\label{sec:data_source}

All experiments used the public EchoNext Mini-Model v1.1.0 release, a PhysioNet dataset containing 100{,}000 de-identified 12-lead ECGs with echocardiography-derived SHD labels \cite{elias2025echonext,goldberger2000physionet}. The public release was derived from routine clinical care at Columbia University Irving Medical Center and includes adults (\( \geq 18 \) years) who had a digitally stored 12-lead ECG and a TTE within a 1-year interval between 2008 and 2022. ECG waveforms were extracted from the GE MUSE management system at 250\,Hz across all 12 leads, and echocardiographic variables were abstracted from Syngo Dynamics (Siemens) and Xcelera (Philips). For supervised experiments, we used the release-defined train, validation, and test partitions. We did not create new patient-level or record-level splits. The training partition may contain multiple ECGs per patient, whereas the validation and test partitions retain only the latest ECG per patient. 

\subsection{Inputs and preprocessing}
\label{sec:inputs_preproc}

Each ECG was analyzed as a fixed 10-second, 12-lead recording sampled at 250\,Hz. To preserve comparability with the public EchoNext benchmark, all waveform-based experiments used the released preprocessed signals. The released waveforms had already undergone lead-wise median filtering, amplitude clipping at the 0.1st and 99.9th percentiles, and normalization using the dataset-wide mean and standard deviation. Each record was represented as a \(12 \times 2500\) lead-by-time matrix. For tabular information, we used the seven ECG-associated covariates distributed with the benchmark, namely sex, ventricular rate, atrial rate, PR interval, QRS duration, corrected QT interval, and age at ECG acquisition. These covariates were used as released, following the public preprocessing protocol.

\subsection{Label construction and task formulation}
\label{sec:labels_targets}

EchoNext labels are derived from structured echocardiography report fields \cite{elias2025echonext,poterucha2025echonext}. The source variables include continuous measurements such as LVEF, interventricular septal thickness, posterior wall thickness, pulmonary artery systolic pressure (PASP), and tricuspid regurgitation maximum velocity, together with categorical assessments of valvular disease, right ventricular (RV) systolic function, and pericardial effusion. For wall thickness, the larger of the interventricular septal (IVS) and left ventricular posterior wall (LVPW) measurements is treated as left ventricular wall thickness (LVWT). In the underlying EchoNext labeling scheme, an ECG is labeled positive for a condition if it occurs within 1 year before an echocardiogram demonstrating that condition. We prespecified six moderate-or-greater endpoints spanning ventricular dysfunction, hypertrophy, and clinically significant valvular disease. These six endpoints were taken directly from the dataset-provided binary flags listed in Table~\ref{tab:targets}.

\begin{table}[!htbp]
\centering
\footnotesize
\setlength{\tabcolsep}{6pt}
\renewcommand{\arraystretch}{1.15}
\begin{tabular}{@{}cc@{}}
\toprule
\textbf{Target} &
\makecell[c]{\textbf{Positive definition in the public release}} \\
\midrule
Reduced LVEF
& LVEF \(\leq 45\%\) \\

Increased LVWT
& \makecell[c]{LVWT \(\geq 1.3\) cm, with LVWT defined as \(\max(\mathrm{IVS}, \mathrm{LVPW})\)} \\

Aortic stenosis
& Moderate or severe aortic stenosis \\

Mitral regurgitation
& Moderate or severe mitral regurgitation \\

Tricuspid regurgitation
& Moderate or severe tricuspid regurgitation \\

RV systolic dysfunction
& Moderate or severe RV systolic dysfunction \\
\bottomrule
\end{tabular}

\caption{Study targets and positive definitions used in the primary benchmark.}
\label{tab:targets}
\end{table}

We formulated the task as a multi-label classification problem because a given ECG could be associated with none, one, or multiple study-defined structural abnormalities. Let \(\mathcal{D}=\{(w_i,u_i,y_i)\}_{i=1}^{N}\) denote the study dataset, where \(w_i \in \mathbb{R}^{12 \times 2500}\) represents the waveform input, \(u_i \in \mathbb{R}^{7}\) denotes the optional covariate vector, and \(y_i \in \{0,1\}^{M}\) denotes the binary target vector for the \(M=6\) study endpoints. For \(j=1,\ldots,M\), \(y_{i,j}=1\) indicates that target \(j\) is present according to the dataset-defined moderate-or-greater criterion. For each record, the model produces a logit vector \(z_i \in \mathbb{R}^{M}\), which is converted to predicted probabilities \(\hat{p}_i \in (0,1)^{M}\) by element-wise application of the sigmoid function,
\(
\hat{p}_{i,j}=\sigma(z_{i,j})\), \( j=1,\ldots,M,\)
where \(\sigma(\cdot)\) denotes the logistic sigmoid. For neural architectures, a waveform encoder \(g_{\theta}(\cdot)\) maps \(w_i\) to a latent representation sequence \(H_i=g_{\theta}(w_i)\), which is subsequently aggregated by a pooling operator to obtain a fixed-dimensional embedding \(h_i=\mathrm{pool}(H_i)\). A prediction head \(h_{\phi}(\cdot)\) then maps \(h_i\), and when applicable a tabular embedding derived from \(u_i\), to the multi-label logit vector \(z_i\). This formulation enables shared representation learning across correlated targets while preserving separate output units for endpoint-specific prediction.

\subsection{Benchmark models}
\label{sec:models}

\subsubsection{Baseline A (engineered features with gradient boosting)}
\label{sec:model_baselineA}

Baseline~A served as a conventional machine-learning benchmark based on engineered ECG descriptors. For each 10-second, 12-lead ECG, a 166-D feature vector was extracted using NeuroKit2 \cite{makowski2021neurokit2}. The resulting feature set summarized beat timing and variability, waveform morphology, band-limited spectral characteristics, and inter-lead relationships. Beat-level measurements were aggregated into record-level descriptors using summary statistics, including means, standard deviations, and robust extrema. Multi-label prediction was implemented in a one-vs-rest manner by training one gradient-boosted classifier per endpoint. Candidate boosting families were first screened under a common validation protocol, after which XGBoost was selected for focused hyperparameter optimization with Optuna \cite{akiba2019optuna,chen2016xgboost}. The six target-specific classifiers jointly produced the probability vector \(\hat{p}_i \in (0,1)^6\); feature-count sensitivity and global importance analyses are summarized in Appendix~\ref{app:baselineA_features}.

\subsubsection{Baseline B (ResNet1D trained from scratch)}
\label{sec:model_baselineB}

Baseline~B was an end-to-end waveform model trained from random initialization. The architecture was derived from ResNet-18 by replacing 2-D convolutions with 1-D temporal convolutions and by treating the 12 ECG leads as input channels \cite{he2016resnet}. The encoder comprised an initial temporal convolution with kernel size 7 and stride 2, followed by max pooling and four residual stages with channel widths of 64, 128, 256, and 512, respectively. Each stage contained two basic residual blocks. When the channel dimension increased, the first block in the stage used stride 2 and the identity branch used a \(1 \times 1\) convolution for dimensional matching. Global average pooling was then applied to obtain a 512-dimensional representation, which was linearly projected to a 768-D embedding and passed to a lightweight multilayer perceptron classifier with dropout to generate the six output logits.

\subsubsection{Baseline C (in-domain adapted ECG-FM with a frozen backbone)}
\label{sec:model_baselineC}

Baseline~C evaluated supervised probing of an in-domain adapted ECG FM. The backbone was ECG-FM, which consists of a convolutional feature extractor followed by a 12-layer transformer encoder with hidden dimension of 768 \cite{mckeen2025ecgfm}. Starting from the released ECG-FM checkpoint, self-supervised pretraining was continued on the EchoNext training waveforms without use of target labels, following the released ECG-FM pretraining framework based on \emph{wav2vec~2.0}-style latent masking and contrastive learning \cite{mckeen2025ecgfm,baevski2020wav2vec}. After this in-domain adaptation stage, all backbone parameters were frozen and only a two-layer multilayer perceptron classifier was trained on mean-pooled token embeddings. Accordingly, this baseline represents an adapted frozen-backbone probing configuration rather than an off-the-shelf frozen encoder without target-domain adaptation.

\subsubsection{ECG-FM with in-domain self-supervised adaptation and partial fine-tuning}
\label{sec:model_proposed}

The primary transfer model used the same in-domain self-supervised adaptation stage as Baseline~C, followed by supervised fine-tuning. We parameterized supervised transformer adaptation by the number of trainable upper transformer blocks, denoted \(b\). For a transformer encoder with \(L=12\) blocks, \(b \in \{1,5,9,12\}\) indicates that the uppermost \(b\) transformer blocks were updated during supervised training, whereas the lower \(L-b\) blocks remained frozen. In the partial fine-tuning family, the convolutional feature extractor remained frozen. Thus, \(b=12\) with the convolutional extractor frozen corresponds to full transformer fine-tuning but not full model fine-tuning. We also evaluated a full-model fine-tuning variant in which both all 12 transformer blocks and the convolutional feature extractor were updated. A separate probing configuration fixed the convolutional extractor and all transformer blocks (\(b=0\)) and trained only the classifier head. All variants used the same pooling operation and classifier-head architecture as Baseline~C.

\subsubsection{Ablations and extensions}
\label{sec:model_ablations}

\textbf{Late fusion with tabular covariates}: To evaluate whether the release-provided covariates contributed information beyond the waveform representation, we performed late-fusion experiments in which the covariate vector \(u_i \in \mathbb{R}^{7}\) was first mapped to an embedding \(e_i \in \mathbb{R}^{d}\) by a multilayer perceptron. The tabular embedding was then combined with the pooled ECG representation \(h_i\) using one of three fusion operators: 1) concatenation, in which \(\tilde{h}_i=[h_i;e_i]\) was passed to a classifier; 2) gated fusion, in which \(\lambda_i=\sigma\!\left(W_g[h_i;e_i]+b_g\right)\) and \(h_i^{\mathrm{fuse}}=\lambda_i \odot h_i + (1-\lambda_i)\odot e_i\); and 3) cross-attention, in which the tabular embedding queried the ECG token sequence before pooling.

\textbf{LoRA parameter-efficient adaptation}: As a parameter-efficient transfer baseline, we evaluated low-rank adaptation (LoRA) \cite{hu2021lora}. In this setting, the ECG-FM backbone weights were frozen and trainable low-rank updates were introduced into the attention projection matrices. For a projection matrix \(W\), the adapted parameterization was defined as
\(
W' = W + BA,
\)
where \(A\) and \(B\) are low-rank matrices of rank \(r\). LoRA modules were applied to the query and value projections in each transformer block, and only the LoRA parameters together with the supervised prediction head were updated during training.

\textbf{Additional ECG FMs}: To examine whether the observed transfer trends extended beyond ECG-FM, we evaluated ECGFounder and HuBERT-ECG as additional open ECG FM backbones \cite{li2025ecgfounder,coppola2024hubertecg}. Each model was assessed under a common downstream protocol using the same EchoNext data partitions, waveform inputs, pooling rule, and multi-label prediction-head family. Backbone-specific implementation adjustments were permitted only when required by architectural differences. Because these encoders are not architecturally identical, these experiments were intended as protocol-aligned comparisons rather than strictly parameter-matched benchmarks.

\textbf{Mixture of FMs (MoFM)}: We further evaluated a MoFM extension in which multiple ECG encoders operated in parallel and their outputs were combined before final prediction. Three fusion strategies were considered: (i) concatenation of projected backbone embeddings, (ii) gated fusion with a softmax gating network that generated sample-specific expert weights, and (iii) logit-level mixture, in which each expert produced its own logits and the final output was formed as a convex combination of the expert predictions. Unless otherwise specified, the encoder-freezing policy for each expert was matched to the corresponding single-backbone configuration.

\textbf{Task-decomposed binary training}: To assess the effect of shared multi-label representation learning, we trained six independent ECG-FM classifiers, one for each study endpoint, using the same backbone family and downstream optimization procedure as in the corresponding multi-label setting. During evaluation, the six target-specific probability estimates were concatenated to form a 6-D prediction vector, and performance was summarized by macro-averaging across endpoints.

\textbf{All-negative downsampling}: To examine the effect of training-set class composition on operating behavior, we conducted a training-only downsampling ablation in which all records containing at least one positive study label were retained, while records with no positive study labels were randomly subsampled according to a retention ratio \(\rho \in \{1.0, 0.5, 0.3, 0.1\}\). The validation and test partitions were left unchanged for all downsampling experiments.

\section{Results}
\label{sec:results}

\subsection{Overall comparison on the test split}
\label{sec:results_overall}

We first evaluated the adapted ECG-FM transfer strategy against three comparator families on the held-out test split. Because the six study targets are imbalanced, we treat macro AUROC and macro AUPRC as the primary discrimination metrics, whereas accuracy and F1 at the fixed threshold \(\tau=0.5\) are reported as secondary operating-point summaries. Table~\ref{tab:main_results} reports macro-averaged test performance across the six SHD targets. Split-wise target prevalence is provided in Appendix~\ref{app:prevalence}, per-label test performance for the high-performing partial-adaptation settings \(b{=}9\) and \(b{=}5\) is reported in Appendix~\ref{app:per_label}, and pairwise target co-occurrence is summarized in Appendix~\ref{app:cooccurrence}.

Among the tested configurations, adapted ECG-FM with \(b{=}9\) trainable upper transformer blocks achieved the highest macro AUROC (0.8509) and accuracy (0.9089), while full transformer fine-tuning with the convolutional feature extractor frozen achieved the highest macro AUPRC (0.4297). A more parameter-efficient setting, \(b{=}5\), yielded nearly identical AUROC (0.8501) with fewer trainable parameters and the highest fixed-threshold F1 (0.3691). Among the comparator baselines, the scratch ResNet1D and the adapted frozen-backbone ECG-FM probe both exceeded engineered XGBoost on the primary discrimination metrics, whereas the frozen-backbone probe achieved the highest F1 among the non-partially-tuned models while updating only 0.60\,M parameters. Overall, the strongest operating points were obtained by supervised adaptation of an in-domain aligned ECG foundation backbone, with different adaptation depths favoring different evaluation priorities.

\begin{table}[!htbp]
\centering
\small
\setlength{\tabcolsep}{4pt}
\resizebox{\linewidth}{!}{%
\begin{tabular}{lccccc}
\toprule
Model & Trainable Params & AUROC & AUPRC & Acc & F1 \\
\midrule
Baseline A: XGBoost & 1.04\,M & 0.8017 & 0.3289 & 0.9012 & 0.1760 \\
Baseline B: ResNet1D & 9.72\,M & 0.8246 & 0.3833 & 0.9063 & 0.1637 \\
Baseline C: adapted ECG-FM probe & 0.60\,M & 0.8233 & 0.3670 & 0.9043 & 0.1933 \\
\midrule
Adapted ECG-FM transformer FT (\(b{=}12\), conv frozen) & 91.08\,M & 0.8501 & \textbf{0.4297} & 0.9081 & 0.3338 \\
Adapted ECG-FM partial FT (\(b{=}9\)) & 69.81\,M & \textbf{0.8509} & 0.4261 & \textbf{0.9089} & 0.3064 \\
Adapted ECG-FM partial FT (\(b{=}5\)) & 41.46\,M & 0.8501 & 0.4173 & 0.9050 & \textbf{0.3691} \\
\bottomrule
\end{tabular}%
}

\caption{Macro-averaged test-set performance across the six SHD targets. Trainable parameter counts reflect parameters updated during the supervised stage only. Bold indicates the best value in each column.}
\label{tab:main_results}
\end{table}

\subsection{Adaptation-depth ablation}
\label{sec:results_freezing}

We next examined how performance changed as progressively larger portions of the in-domain adapted ECG-FM backbone were updated during supervised training. Table~\ref{tab:freezing_depth} reports the fully frozen probe (\(b{=}0\)), partial fine-tuning with \(b \in \{1,5,9\}\) trainable upper transformer blocks, full transformer fine-tuning (\(b{=}12\)) with the convolutional feature extractor frozen, and full model fine-tuning where both the convolutional extractor and all transformer blocks were updated. This distinction is important because \(b\) denotes the number of trainable upper transformer blocks, whereas the convolutional feature extractor is handled separately. Performance did not improve monotonically with trainable parameter count. Full transformer fine-tuning with the convolutional extractor frozen achieved the highest macro AUPRC (0.4297), \(b{=}9\) achieved the highest macro AUROC (0.8509) and accuracy (0.9089), and \(b{=}5\) preserved nearly identical AUROC with fewer trainable parameters and produced the highest fixed-threshold F1 (0.3691). Full model fine-tuning did not improve over full transformer fine-tuning, suggesting that updating the convolutional extractor was unnecessary. Shallow adaptation (\(b{=}1\)) remained stronger than the frozen probe on AUROC and AUPRC, but was clearly weaker than deeper partial fine-tuning.

\begin{table}[!htbp]
\centering
\small
\setlength{\tabcolsep}{4pt}
\resizebox{\linewidth}{!}{%
\begin{tabular}{lccccc}
\toprule
Supervised adaptation setting & Trainable Params & AUROC & AUPRC & Acc & F1 \\
\midrule
Full model FT (\(b{=}12\), conv unfrozen) & 91.48\,M & 0.8484 & 0.4198 & 0.9041 & 0.3423 \\
Full transformer FT (\(b{=}12\), conv frozen) & 91.08\,M & 0.8501 & \textbf{0.4297} & 0.9081 & 0.3338 \\
Partial FT (\(b{=}9\), conv frozen) & 69.81\,M & \textbf{0.8509} & 0.4261 & \textbf{0.9089} & 0.3064 \\
Partial FT (\(b{=}5\), conv frozen) & 41.46\,M & 0.8501 & 0.4173 & 0.9050 & \textbf{0.3691} \\
Partial FT (\(b{=}1\), conv frozen) & 13.11\,M & 0.8398 & 0.3977 & 0.9078 & 0.2381 \\
Frozen backbone probe (\(b{=}0\)) & 0.60\,M & 0.8233 & 0.3670 & 0.9043 & 0.1933 \\
\bottomrule
\end{tabular}%
}

\caption{Effect of supervised adaptation depth on macro-averaged test performance across the six SHD targets. \(b\) denotes the number of trainable upper transformer blocks.}
\label{tab:freezing_depth}
\end{table}

\subsection{Performance--efficiency operating points}
\label{sec:results_tradeoff}

\begin{figure}[!htbp]
    \centering    \includegraphics[width=\linewidth]{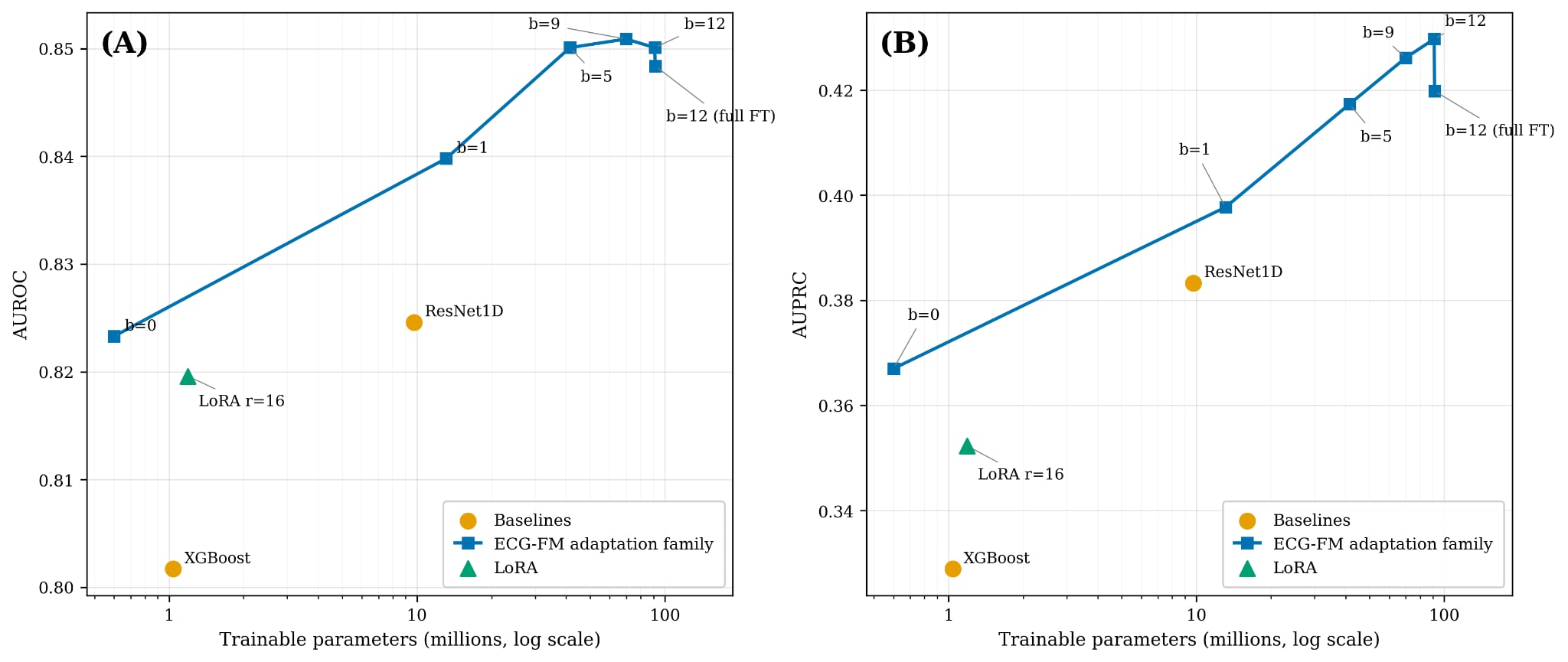}

    \caption{Performance--efficiency operating points for the main benchmark models and ECG-FM adaptation settings. The x-axis shows trainable parameters on a log scale. \textbf{(A)} AUROC vs.\ trainable parameters. \textbf{(B)} AUPRC vs.\ trainable parameters.}
    \label{fig:perf_efficiency}
\end{figure}

Because the compared models span substantially different parameter regimes, Figure~\ref{fig:perf_efficiency} summarizes their performance--efficiency operating points on a log-parameter scale. This view is intended as an operating-point summary rather than a parameter-matched benchmark. Across the ECG-FM adaptation family, moving from the frozen probe to supervised backbone adaptation produces large gains in AUROC and AUPRC. However, performance saturates after strong partial or full-transformer fine-tuning: \(b{=}9\) achieves the best AUROC, \(b{=}12\) with the convolutional extractor frozen achieves the best AUPRC, and full model fine-tuning does not improve over these operating points despite updating more parameters.

\subsection{Fusion and LoRA ablations}
\label{sec:results_fusion_lora}

We next asked whether the adapted waveform backbone benefited from adding the seven release-provided tabular covariates and whether parameter-efficient adaptation could approach partial fine-tuning with far fewer trainable parameters. Because AUROC and AUPRC are the primary discrimination metrics in this imbalanced multi-label setting, we use the \(b{=}9\) adapted ECG-FM configuration as the waveform-only anchor because it achieved the highest AUROC and remained one of the strongest threshold-independent operating points in Table~\ref{tab:freezing_depth}.

Table~\ref{tab:fusion_lora} shows that none of the tested fusion operators improved AUROC or AUPRC over the waveform-only \(b{=}9\) baseline. Cross-attention yielded slightly higher accuracy (0.9093), and gated fusion produced the highest F1 among the fusion variants (0.3324), but both came with lower AUROC and AUPRC than waveform-only transfer. The tested LoRA configuration reduced the supervised trainable-parameter count to 1.19\,M, but it did not approach partial fine-tuning on either AUROC or AUPRC. The added tabular covariates did not provide measurable threshold-independent benefit beyond the adapted waveform representation, and the evaluated LoRA setting did not match selective weight updates.

\begin{table}[!htbp]
\centering
\resizebox{\linewidth}{!}{%
\begin{tabular}{lccccc}
\toprule
Method & Trainable Params & AUROC & AUPRC & Acc & F1 \\
\midrule
Adapted ECG-FM (\(b{=}9\), waveform-only) & 69.81\,M & \textbf{0.8509} & \textbf{0.4261} & 0.9089 & 0.3064 \\
\midrule
Cross-attention fusion & 69.81\,M & 0.8487 & 0.4167 & \textbf{0.9093} & 0.3140 \\
Concat fusion & 69.74\,M & 0.8478 & 0.4134 & 0.9083 & 0.2878 \\
Gated fusion & 69.61\,M & 0.8469 & 0.4195 & 0.9071 & \textbf{0.3324} \\
\midrule
LoRA (rank \(=16\)) & 1.19\,M & 0.8196 & 0.3523 & 0.9045 & 0.1963 \\
\bottomrule
\end{tabular}%
}

\caption{Ablation results for late fusion and LoRA. Fusion models use the adapted ECG-FM \(b{=}9\) backbone and incorporate the seven release-provided covariates through the stated fusion operator. Metrics are macro-averaged across the six SHD targets.}
\label{tab:fusion_lora}
\end{table}

\subsection{Additional experiments: foundation backbones, mixtures, task decomposition, and downsampling}
\label{sec:results_extensions}

Table~\ref{tab:foundation_model_comparison} summarizes results for the tested configurations of three open ECG FM backbones under a common downstream evaluation pipeline. Because these backbones differ in architecture and were not all evaluated under identical supervised adaptation settings, the comparison should be interpreted as protocol-aligned instead of a strictly parameter-matched head-to-head benchmark. Within this comparison, in-domain adapted ECG-FM with \(b{=}5\) achieved the numerically highest macro AUROC (0.8501), macro AUPRC (0.4173), and fixed-threshold F1 (0.3691). Holding the supervised configuration constant, removing the EchoNext self-supervised adaptation stage from ECG-FM reduced AUROC to 0.8387, AUPRC to 0.4048, and F1 to 0.3166. Accuracy varied only modestly across the tested configurations (0.9021--0.9058), indicating that the main separation among backbones appeared in ranking quality and precision--recall behavior rather than in thresholded accuracy alone. A qualitative embedding visualization is provided in Appendix~\ref{app:tsne}, but it is descriptive only and is not used to support model-ranking claims.

\begin{table}[!htbp]
\centering
\small
\setlength{\tabcolsep}{4pt}
\resizebox{\linewidth}{!}{%
\begin{tabular}{llccccc}
\toprule
Model & Tested configuration & Trainable Params & AUROC & AUPRC & Acc & F1 \\
\midrule
ECG-FM (in-domain adapted) & partial FT (\(b{=}5\)) & 41.46\,M & \textbf{0.8501} & \textbf{0.4173} & 0.9050 & \textbf{0.3691} \\
ECG-FM (no in-domain adaptation) & partial FT (\(b{=}5\)) & 41.46\,M & 0.8387 & 0.4048 & \textbf{0.9058} & 0.3166 \\
ECGFounder & full fine-tuning & 31.46\,M & 0.8228 & 0.3744 & 0.9043 & 0.2290 \\
HuBERT-ECG & partial FT (\(b{=}5\)) & 36.43\,M & 0.8182 & 0.3776 & 0.9021 & 0.2949 \\
\bottomrule
\end{tabular}%
}

\caption{Comparison of open ECG FM backbones under a common downstream evaluation pipeline. For ECG-FM, we additionally report results without Stage~1 in-domain self-supervised adaptation on EchoNext.}
\label{tab:foundation_model_comparison}
\end{table}

Table~\ref{tab:mofm} evaluates MoFM variants that combine ECG-FM, ECGFounder, and HuBERT-ECG using three fusion strategies. We use the adapted ECG-FM \(b{=}5\) model as the single-backbone reference because it provided the strongest fixed-threshold F1 with a smaller trainable-parameter budget. The MoFM variants produced slightly higher macro AUPRC (0.4184--0.4194 versus 0.4173) and accuracy (0.9082--0.9085 versus 0.9050), but all reduced macro AUROC (0.8461--0.8481 versus 0.8501) and fixed-threshold F1 (0.2997--0.3253 versus 0.3691), while increasing the trainable-parameter budget to approximately 105--106\,M.

\begin{table}[!htbp]
\centering
\resizebox{\linewidth}{!}{%
\begin{tabular}{lccccc}
\toprule
Method & Trainable Params & AUROC & AUPRC & Acc & F1 \\
\midrule
Adapted ECG-FM (\(b{=}5\), single model) & 41.46\,M & \textbf{0.8501} & 0.4173 & 0.9050 & \textbf{0.3691} \\
\midrule
MoFM: Concat fusion & $\sim$106\,M & 0.8461 & 0.4192 & \textbf{0.9085} & 0.3210 \\
MoFM: Gated fusion & $\sim$106\,M & 0.8478 & \textbf{0.4194} & \textbf{0.9085} & 0.2997 \\
MoFM: Logit MoE & $\sim$105\,M & 0.8481 & 0.4184 & 0.9082 & 0.3253 \\
\bottomrule
\end{tabular}%
}

\caption{Comparison of the adapted single-backbone ECG-FM baseline and MoFM fusion strategies combining ECG-FM, ECGFounder, and HuBERT-ECG.}
\label{tab:mofm}
\end{table}

Table~\ref{tab:binary_vs_multilabel} compares the unified multi-label ECG-FM model with an ensemble of six independently trained binary ECG-FM models under the tested binary configuration. The unified multi-label model achieved higher macro AUROC (0.8501 versus 0.8459), higher macro AUPRC (0.4173 versus 0.4111), and substantially higher fixed-threshold F1 (0.3691 versus 0.2769), while using a much smaller total trainable-parameter budget than the six-model binary ensemble. The binary ensemble achieved slightly higher accuracy (0.9076 versus 0.9050), but this did not translate into better discrimination or precision--recall behavior. Because the binary ensemble used a different supervised adaptation depth from the unified model, this should be interpreted as a comparison of tested configurations rather than as a pure decomposition-only ablation.

\begin{table}[!htbp]
\centering
\resizebox{\linewidth}{!}{%
\begin{tabular}{lccccc}
\toprule
Approach & Trainable Params & AUROC & AUPRC & Acc & F1 \\
\midrule
ECG-FM multi-label (\(b{=}5\)) & 41.46\,M & \textbf{0.8501} & \textbf{0.4173} & 0.9050 & \textbf{0.3691} \\
\(6\times\) ECG-FM binary (\(b{=}3\) each) & \(6 \times 27.28\)\,M & 0.8459 & 0.4111 & \textbf{0.9076} & 0.2769 \\
\bottomrule
\end{tabular}%
}

\caption{Comparison of a unified multi-label model against an ensemble of six task-specific binary classifiers. The reported parameter count for the binary approach is the total across all six independently trained models.}
\label{tab:binary_vs_multilabel}
\end{table}

Table~\ref{tab:downsampling} reports the effect of random downsampling of all-negative training records for adapted ECG-FM with \(b{=}5\), while leaving the validation and test splits unchanged. Downsampling did not improve threshold-independent discrimination monotonically. Relative to the unmodified training subset (\(\rho{=}1.0\)), light downsampling (\(\rho{=}0.5\)) yielded a very similar macro AUPRC (0.4190 versus 0.4173) with slightly lower AUROC (0.8467 versus 0.8501), whereas more aggressive downsampling progressively reduced both AUROC and AUPRC. By contrast, fixed-threshold F1 increased most at \(\rho{=}0.3\), reaching 0.4103. This pattern suggests that all-negative downsampling primarily changes score calibration and operating behavior at \(\tau=0.5\). We therefore interpret this ablation as a pragmatic operating-point manipulation rather than as evidence of a better classifier in a threshold-independent sense.

\begin{table}[!htbp]
\centering
\resizebox{\linewidth}{!}{%
\begin{tabular}{lcccccc}
\toprule
Retention Ratio \(\rho\) & Records Removed & Train Set Size & AUROC & AUPRC & Acc & F1 \\
\midrule
1.0 (no downsampling) & 0 & 72{,}297 & \textbf{0.8501} & 0.4173 & \textbf{0.9050} & 0.3691 \\
0.5 & 19{,}016 & 53{,}459 & 0.8467 & \textbf{0.4190} & 0.9026 & 0.3618 \\
0.3 & 26{,}622 & 45{,}853 & 0.8415 & 0.4050 & 0.8985 & \textbf{0.4103} \\
0.1 & 34{,}228 & 38{,}247 & 0.8242 & 0.3830 & 0.8739 & 0.3765 \\
\bottomrule
\end{tabular}%
}

\caption{Effect of all-negative training-record downsampling on adapted ECG-FM with \(b{=}5\). Here \(\rho{=}1.0\) denotes the unmodified training subset used for this ablation.}
\label{tab:downsampling}
\end{table}

\section{Discussion}
\label{sec:discussion}

The adaptation-depth results suggest that transfer quality depended not only on the presence of large-scale pretraining, but also on how much of the pretrained representation was allowed to specialize to the EchoNext distribution. The observation that the same ECG-FM backbone improved after continued self-supervised training on EchoNext, and that the best partial fine-tuning settings outperformed both the adapted frozen probe and full model fine-tuning, supports the view that target-domain alignment and selective adaptation can provide a favorable performance--efficiency operating point. This interpretation is consistent with the broader motivation of ECG FMs, which aim to provide transferable representations that can reduce labeling burden and improve downstream generalization \cite{mckeen2025ecgfm}. At the same time, the cross-backbone comparison should be interpreted cautiously. Although adapted ECG-FM achieved the strongest performance among the tested configurations, the evaluated backbones are not architecturally identical and were not all assessed under perfectly parameter-matched adaptation settings.

The ablation experiments further clarify which extensions were and were not beneficial under the present protocol. First, late fusion of the seven release-provided covariates did not improve AUROC or AUPRC relative to waveform-only partial fine-tuning, suggesting that the adapted waveform representation already captured much of the predictive signal available from those variables. Second, the evaluated LoRA configuration substantially reduced the number of supervised trainable parameters, but it did not approach selective partial fine-tuning on the primary discrimination metrics. Third, MoFM variants did not produce a better overall performance--efficiency operating point than the strongest adapted single-backbone model, despite modest shifts in the precision--recall trade-off for some fusion schemes. Fourth, the unified multi-label model outperformed the tested ensemble of task-specific binary models, which is compatible with the hypothesis that shared representation learning can be advantageous when related structural abnormalities co-occur and must be modeled jointly. Finally, all-negative downsampling mainly altered fixed-threshold behavior rather than uniformly improving threshold-independent discrimination, indicating that its primary effect was to shift the operating point rather than to produce a better classifier.

From a clinical perspective, these findings are most appropriately interpreted as evidence for ECG-based case finding or echocardiography triage. The EchoNext labeling framework links each ECG to moderate-or-greater echocardiographic abnormalities identified within the subsequent year, thereby making the task clinically relevant for referral support while also defining the prediction target through a specific temporal and report-derived labeling rule \cite{poterucha2025echonext}. In addition, the per-label results indicate heterogeneous difficulty across endpoints, with stronger performance for some ventricular dysfunction targets than for certain valvular targets. Macro-averaged performance is useful for benchmarking, but real-world deployment would require endpoint-specific operating thresholds, explicit consideration of the relative consequences of false positives and false negatives, and calibration analyses to determine whether predicted probabilities are sufficiently reliable for clinical decision support.

Several limitations should be acknowledged. First, the study used a public benchmark derived from a single health system, and the mini-model release does not substitute for independent external validation of the specific transfer strategy studied here, even though the full EchoNext study reported external validation of the original EchoNext model across multiple care settings \cite{poterucha2025echonext}. Second, the labels are derived from structured echocardiography reports and a one-year ECG--echo linkage rule, which may introduce temporal mismatch and residual label noise for some records. Finally, the comparison across open foundation backbones was protocol-aligned but not strictly parameter-matched, and the binary-ensemble comparison used a different adaptation depth from the unified multi-label model.

\section{Conclusion}
\label{sec:conclusion}

In conclusion, this study shows that open pretrained ECG models can support echo-confirmed multi-label SHD detection when adapted in a targeted and computationally practical manner. On the EchoNext Mini-Model benchmark, continued in-domain self-supervised adaptation followed by selective supervised fine-tuning of ECG-FM provided the most favorable overall performance among the evaluated configurations, outperforming engineered-feature and from-scratch waveform baselines while avoiding the unnecessary cost of broader end-to-end updating. More broadly, the results support open ECG FMs as a practical and reproducible basis for SHD detection from routine 12-lead ECGs. Future work should extend this benchmark-centered analysis toward external multi-institutional validation, probability calibration and clinically specified threshold selection, subgroup robustness assessment, and prospective workflow evaluation to determine whether the observed discrimination gains translate into reliable and clinically actionable decision support in real-world practice.

\printbibliography

@article{siontis2021aiecg,
  author  = {Siontis, Konstantinos C. and Noseworthy, Peter A. and Attia, Zachi I. and Friedman, Paul A.},
  title   = {Artificial intelligence-enhanced electrocardiography in cardiovascular disease management},
  journal = {Nature Reviews Cardiology},
  year    = {2021},
  volume  = {18},
  number  = {7},
  pages   = {465--478},
  doi     = {10.1038/s41569-020-00503-2},
  url     = {https://www.nature.com/articles/s41569-020-00503-2}
}

@article{ulloa2022rechommend,
  author  = {Ulloa-Cerna, Alvaro E. and others},
  title   = {{rECHOmmend}: An {ECG}-Based Machine Learning Approach for Identifying Patients at High Risk of Undiagnosed Structural Heart Disease Detectable by Echocardiography},
  journal = {Circulation},
  year    = {2022},
  volume  = {146},
  number  = {1},
  pages   = {36--47},
  doi     = {10.1161/CIRCULATIONAHA.121.057869},
  url     = {https://www.ahajournals.org/doi/10.1161/CIRCULATIONAHA.121.057869}
}

@article{poterucha2025echonext,
  author  = {Poterucha, Timothy J. and Jing, Linyuan and Ricart, Ramon Pimentel and others},
  title   = {Detecting structural heart disease from electrocardiograms using {AI}},
  journal = {Nature},
  year    = {2025},
  volume  = {644},
  number  = {8075},
  pages   = {221--230},
  doi     = {10.1038/s41586-025-09227-0},
  url     = {https://www.nature.com/articles/s41586-025-09227-0}
}

@article{hannun2019arrhythmia,
  author  = {Hannun, Awni Y. and Rajpurkar, Pranav and Haghpanahi, Masoumeh and Tison, Geoffrey H. and Bourn, Clare and Turakhia, Mintu P. and Ng, Andrew Y.},
  title   = {Cardiologist-level arrhythmia detection and classification in ambulatory electrocardiograms using a deep neural network},
  journal = {Nature Medicine},
  year    = {2019},
  volume  = {25},
  number  = {1},
  pages   = {65--69},
  doi     = {10.1038/s41591-018-0268-3},
  url     = {https://www.nature.com/articles/s41591-018-0268-3}
}

@article{ribeiro2020automatic,
  author  = {Ribeiro, Ant{\^o}nio H. and Ribeiro, Manoel Horta and Paix{\~a}o, Gabriela M. M. and others},
  title   = {Automatic diagnosis of the 12-lead {ECG} using a deep neural network},
  journal = {Nature Communications},
  year    = {2020},
  volume  = {11},
  pages   = {1760},
  doi     = {10.1038/s41467-020-15432-4},
  url     = {https://www.nature.com/articles/s41467-020-15432-4}
}

@article{wagner2020ptbxl,
  author  = {Wagner, Patrick and Strodthoff, Nils and Bousseljot, Ralf-Dieter and Kreiseler, Dieter and Lunze, Fatima I. and Samek, Wojciech and Schaeffter, Tobias and others},
  title   = {{PTB-XL}, a large publicly available electrocardiography dataset},
  journal = {Scientific Data},
  year    = {2020},
  volume  = {7},
  pages   = {154},
  doi     = {10.1038/s41597-020-0495-6},
  url     = {https://www.nature.com/articles/s41597-020-0495-6}
}

@article{strodthoff2021ptbxl,
  author  = {Strodthoff, Nils and Wagner, Patrick and Schaeffter, Tobias and Samek, Wojciech},
  title   = {Deep Learning for {ECG} Analysis: Benchmarks and Insights from {PTB-XL}},
  journal = {IEEE Journal of Biomedical and Health Informatics},
  year    = {2021},
  volume  = {25},
  number  = {5},
  pages   = {1519--1528},
  doi     = {10.1109/JBHI.2020.3022989},
  url     = {https://pubmed.ncbi.nlm.nih.gov/32903191/}
}

@article{attia2019lvef,
  author  = {Attia, Zachi I. and Kapa, Suraj and Lopez-Jimenez, Francisco and McKie, Paul M. and others},
  title   = {Screening for cardiac contractile dysfunction using an artificial intelligence-enabled electrocardiogram},
  journal = {Nature Medicine},
  year    = {2019},
  volume  = {25},
  number  = {1},
  pages   = {70--74},
  doi     = {10.1038/s41591-018-0240-2},
  url     = {https://www.nature.com/articles/s41591-018-0240-2}
}

@article{elias2022valvular,
  author  = {Elias, Patrick and others},
  title   = {Deep learning electrocardiographic analysis for detection of left-sided valvular heart disease},
  journal = {Journal of the American College of Cardiology},
  year    = {2022},
  volume  = {80},
  number  = {6},
  pages   = {613--626},
  doi     = {10.1016/j.jacc.2022.05.029},
  url     = {https://www.sciencedirect.com/science/article/pii/S0735109722052251}
}

@article{elias2025echonext,
  author  = {Elias, Pierre and Finer, Joshua},
  title   = {{EchoNext}: A Dataset for Detecting Echocardiogram-Confirmed Structural Heart Disease from {ECG}s},
  journal = {PhysioNet},
  year    = {2025},
  month   = sep,
  note    = {Version 1.1.0},
  doi     = {10.13026/3ykd-bf14},
  url     = {https://doi.org/10.13026/3ykd-bf14}
}

@article{goldberger2000physionet,
  author  = {Goldberger, Ary L. and Amaral, Luis A. N. and Glass, Leon and Hausdorff, Jeffrey M. and Ivanov, Plamen Ch. and Mark, Roger G. and Mietus, Joseph E. and Moody, George B. and Peng, Chung-Kang and Stanley, H. Eugene},
  title   = {{PhysioBank, PhysioToolkit, and PhysioNet}: Components of a New Research Resource for Complex Physiologic Signals},
  journal = {Circulation},
  year    = {2000},
  volume  = {101},
  number  = {23},
  pages   = {e215--e220},
  doi     = {10.1161/01.CIR.101.23.e215},
  url     = {https://pubmed.ncbi.nlm.nih.gov/10851218/}
}

@inproceedings{kiyasseh2021clocs,
  author    = {Kiyasseh, Dani and Zhu, Tingting and Clifton, David A.},
  title     = {{CLOCS}: Contrastive Learning of Cardiac Signals Across Space, Time, and Patients},
  booktitle = {Proceedings of the 38th International Conference on Machine Learning},
  series    = {Proceedings of Machine Learning Research},
  volume    = {139},
  pages     = {5606--5615},
  year      = {2021},
  publisher = {PMLR},
  url       = {https://proceedings.mlr.press/v139/kiyasseh21a.html}
}

@article{mckeen2025ecgfm,
  author  = {McKeen, Kaden and Masood, Sameer and Toma, Augustin and Rubin, Barry and Wang, Bo},
  title   = {{ECG-FM}: an open electrocardiogram foundation model},
  journal = {JAMIA Open},
  year    = {2025},
  volume  = {8},
  number  = {5},
  pages   = {ooaf122},
  doi     = {10.1093/jamiaopen/ooaf122},
  url     = {https://academic.oup.com/jamiaopen/article/doi/10.1093/jamiaopen/ooaf122/8287827}
}

@article{li2025ecgfounder,
  author  = {Li, Jun and Aguirre, Aaron D. and Moura Junior, Valdery and Jin, Jiarui and Liu, Che and Zhong, Lanhai and Sun, Chenxi and Clifford, Gari and Westover, M. Brandon and Hong, Shenda},
  title   = {An Electrocardiogram Foundation Model Built on over 10 Million Recordings with External Evaluation across Multiple Domains},
  journal = {NEJM AI},
  year    = {2025},
  volume  = {2},
  number  = {7},
  pages   = {AIoa2401033},
  doi     = {10.1056/AIoa2401033},
  url     = {https://pubmed.ncbi.nlm.nih.gov/40771651/}
}

@article{coppola2024hubertecg,
  author  = {Coppola, Edoardo and Savardi, Mattia and Massussi, Mauro and Adamo, Marianna and Metra, Marco and Signoroni, Alberto},
  title   = {{HuBERT-ECG}: A Self-Supervised Foundation Model for Broad and Scalable Cardiac Applications},
  journal = {medRxiv},
  year    = {2024},
  doi     = {10.1101/2024.11.14.24317328},
  url     = {https://www.medrxiv.org/content/10.1101/2024.11.14.24317328}
}

@article{makowski2021neurokit2,
  author  = {Makowski, Dominique and Pham, Tam and Lau, Zen J. and Brammer, Jan C. and Lespinasse, Fran{\c{c}}ois and Pham, Hung and Sch{\"o}lzel, Christopher and Chen, S. H. Annabel},
  title   = {{NeuroKit2}: A Python Toolbox for Neurophysiological Signal Processing},
  journal = {Behavior Research Methods},
  year    = {2021},
  volume  = {53},
  number  = {4},
  pages   = {1689--1696},
  doi     = {10.3758/s13428-020-01516-y},
  url     = {https://pubmed.ncbi.nlm.nih.gov/33528817/}
}

@inproceedings{akiba2019optuna,
  author    = {Akiba, Takuya and Sano, Shotaro and Yanase, Toshihiko and Ohta, Takeru and Koyama, Masanori},
  title     = {Optuna: A Next-generation Hyperparameter Optimization Framework},
  booktitle = {Proceedings of the 25th ACM SIGKDD International Conference on Knowledge Discovery and Data Mining},
  year      = {2019},
  pages     = {2623--2631},
  doi       = {10.1145/3292500.3330701},
  url       = {https://dl.acm.org/doi/10.1145/3292500.3330701}
}

@inproceedings{chen2016xgboost,
  author    = {Chen, Tianqi and Guestrin, Carlos},
  title     = {{XGBoost}: A Scalable Tree Boosting System},
  booktitle = {Proceedings of the 22nd ACM SIGKDD International Conference on Knowledge Discovery and Data Mining},
  year      = {2016},
  pages     = {785--794},
  doi       = {10.1145/2939672.2939785},
  url       = {https://dl.acm.org/doi/10.1145/2939672.2939785}
}

@inproceedings{he2016resnet,
  author    = {He, Kaiming and Zhang, Xiangyu and Ren, Shaoqing and Sun, Jian},
  title     = {Deep Residual Learning for Image Recognition},
  booktitle = {Proceedings of the IEEE Conference on Computer Vision and Pattern Recognition (CVPR)},
  year      = {2016},
  pages     = {770--778},
  doi       = {10.1109/CVPR.2016.90},
  url       = {https://doi.org/10.1109/CVPR.2016.90}
}

@misc{hu2021lora,
  author       = {Hu, Edward J. and Shen, Yelong and Wallis, Phillip and Allen-Zhu, Zeyuan and Li, Yuanzhi and Wang, Shean and Wang, Lu and Chen, Weizhu},
  title        = {LoRA: Low-Rank Adaptation of Large Language Models},
  year         = {2021},
  eprint       = {2106.09685},
  archivePrefix= {arXiv},
  primaryClass = {cs.CL},
  doi          = {10.48550/arXiv.2106.09685},
  url          = {https://arxiv.org/abs/2106.09685}
}

@inproceedings{baevski2020wav2vec,
  author    = {Baevski, Alexei and Zhou, Henry and Mohamed, Abdelrahman and Auli, Michael},
  title     = {wav2vec 2.0: A Framework for Self-Supervised Learning of Speech Representations},
  booktitle = {Advances in Neural Information Processing Systems},
  year      = {2020},
  volume    = {33},
  pages     = {12449--12460},
  doi       = {10.48550/arXiv.2006.11477},
  url       = {https://arxiv.org/abs/2006.11477}
}

\clearpage
\appendix
\titleformat{\section}{\large\bfseries}{Appendix \thesection.}{0.5em}{}

\section{Target prevalence in the release cohort and supervised splits}
\label{app:prevalence}

Table~\ref{tab:appendix_prevalence} summarizes the prevalence of the six study endpoints in the full EchoNext release and in the train, validation, and test partitions used for supervised model development and evaluation. These prevalences provide important context for the interpretation of precision--recall performance and fixed-threshold F1, particularly for lower-prevalence endpoints such as aortic stenosis.

\begin{table}[!htbp]
\centering
\small
\begin{tabular}{lcccc}
\toprule
Target & Overall (\%) & Train (\%) & Val (\%) & Test (\%) \\
\midrule
Reduced LVEF (\(\leq 45\%\)) & 23.89 & 23.40 & 18.72 & 17.68 \\
Increased LVWT (\(\geq 1.3\) cm) & 24.22 & 24.38 & 18.96 & 19.50 \\
Aortic stenosis (moderate or greater) & 4.05 & 4.03 & 5.45 & 5.26 \\
Mitral regurgitation (moderate or greater) & 8.45 & 8.47 & 6.10 & 6.19 \\
Tricuspid regurgitation (moderate or greater) & 10.65 & 10.63 & 6.59 & 6.49 \\
RV systolic dysfunction (moderate or greater) & 13.24 & 13.24 & 7.96 & 7.70 \\
\bottomrule
\end{tabular}

\caption{Prevalence of the six study endpoints in the EchoNext release. The Overall column is calculated over the full release cohort, whereas the Train, Val, and Test columns are calculated within the supervised partitions used in this study. The supervised split sizes are train \(n=72{,}475\), validation \(n=4{,}626\), and test \(n=5{,}442\).}
\label{tab:appendix_prevalence}
\end{table}

\section{Test performance for selected partial-adaptation settings}
\label{app:per_label}

Table~\ref{tab:appendix_per_label} reports per-endpoint test performance for two high-performing ECG-FM partial fine-tuning configurations retained in the main analysis. The \(b=9\) setting achieved the highest macro AUROC among partial-adaptation settings, whereas \(b=5\) yielded nearly identical AUROC with fewer trainable parameters and the highest fixed-threshold macro F1. Test-set endpoint prevalence is included to contextualize AUPRC and F1 under class imbalance.

\begin{table}[!htbp]
\centering
\scriptsize
\resizebox{\linewidth}{!}{%
\begin{tabular}{lccccccccc}
\toprule
& & \multicolumn{4}{c}{Adapted ECG-FM partial FT (\(b=9\))} & \multicolumn{4}{c}{Adapted ECG-FM partial FT (\(b=5\))} \\
\cmidrule(lr){3-6}\cmidrule(lr){7-10}
Endpoint & Test Prev. (\%) & AUROC & AUPRC & Acc & F1 & AUROC & AUPRC & Acc & F1 \\
\midrule
Reduced LVEF (\(\leq 45\%\)) & 17.68 & 0.9041 & 0.7425 & 0.8907 & 0.6582 & 0.9015 & 0.7393 & 0.8923 & 0.6667 \\
Increased LVWT (\(\geq 1.3\) cm) & 19.50 & 0.7668 & 0.4188 & 0.8102 & 0.2365 & 0.7648 & 0.4208 & 0.8049 & 0.3461 \\
Aortic stenosis & 5.26 & 0.8547 & 0.2801 & 0.9473 & 0.1681 & 0.8512 & 0.2617 & 0.9434 & 0.1630 \\
Mitral regurgitation & 6.19 & 0.8450 & 0.2788 & 0.9364 & 0.1307 & 0.8362 & 0.2571 & 0.9309 & 0.2134 \\
Tricuspid regurgitation & 6.49 & 0.8515 & 0.3470 & 0.9372 & 0.2083 & 0.8478 & 0.3194 & 0.9267 & 0.3339 \\
RV systolic dysfunction & 7.70 & 0.8831 & 0.4895 & 0.9316 & 0.4364 & 0.8990 & 0.5056 & 0.9316 & 0.4918 \\
\midrule
Macro average & --- & 0.8509 & 0.4261 & 0.9089 & 0.3064 & 0.8501 & 0.4173 & 0.9050 & 0.3691 \\
\bottomrule
\end{tabular}%
}

\caption{Per-endpoint test performance for selected adapted ECG-FM partial fine-tuning configurations. Endpoint-specific test prevalence is shown to contextualize the precision--recall and fixed-threshold results.}
\label{tab:appendix_per_label}
\end{table}

\section{Pairwise co-occurrence structure among the six study endpoints}
\label{app:cooccurrence}

Table~\ref{tab:appendix_cooccurrence} summarizes pairwise co-occurrence among the six modeled endpoints in the full EchoNext release. For each pair, the table reports the joint count, the joint prevalence, and the conditional prevalences in both directions. These quantities provide additional context for the multi-label formulation by indicating which endpoints most frequently co-occur in the release cohort.

\begin{table}[!htbp]
\centering
\scriptsize
\resizebox{\linewidth}{!}{%
\begin{tabular}{llcccc}
\toprule
Endpoint 1 & Endpoint 2 & Count & Joint Prev. (\%) & \(P(\mathrm{T2}=1 \mid \mathrm{T1}=1)\) (\%) & \(P(\mathrm{T1}=1 \mid \mathrm{T2}=1)\) (\%) \\
\midrule
Reduced LVEF & RV systolic dysfunction & 8{,}993 & 8.99 & 37.64 & 67.91 \\
Reduced LVEF & Increased LVWT & 8{,}038 & 8.04 & 33.64 & 33.19 \\
Reduced LVEF & Mitral regurgitation & 5{,}693 & 5.69 & 23.83 & 67.36 \\
Increased LVWT & RV systolic dysfunction & 5{,}312 & 5.31 & 21.93 & 40.11 \\
Tricuspid regurgitation & RV systolic dysfunction & 5{,}169 & 5.17 & 48.53 & 39.03 \\
Reduced LVEF & Tricuspid regurgitation & 5{,}165 & 5.16 & 21.62 & 48.49 \\
Increased LVWT & Tricuspid regurgitation & 3{,}812 & 3.81 & 15.74 & 35.79 \\
Mitral regurgitation & RV systolic dysfunction & 3{,}500 & 3.50 & 41.42 & 26.43 \\
Mitral regurgitation & Tricuspid regurgitation & 3{,}463 & 3.46 & 40.98 & 32.51 \\
Increased LVWT & Mitral regurgitation & 3{,}276 & 3.28 & 13.53 & 38.76 \\
Increased LVWT & Aortic stenosis & 1{,}790 & 1.79 & 7.39 & 44.15 \\
Reduced LVEF & Aortic stenosis & 1{,}327 & 1.33 & 5.55 & 32.73 \\
Aortic stenosis & Mitral regurgitation & 813 & 0.81 & 20.05 & 9.62 \\
Aortic stenosis & Tricuspid regurgitation & 710 & 0.71 & 17.51 & 6.67 \\
Aortic stenosis & RV systolic dysfunction & 551 & 0.55 & 13.59 & 4.16 \\
\bottomrule
\end{tabular}%
}

\caption{Pairwise co-occurrence structure among the six modeled endpoints in the full EchoNext release (\(n=100{,}000\) ECGs). Pairs are ordered by decreasing joint prevalence.}
\label{tab:appendix_cooccurrence}
\end{table}

\section{Engineered-feature sensitivity and importance for Baseline A}
\label{app:baselineA_features}

\begin{figure}[!htbp]
    \centering
    \includegraphics[width=\linewidth]{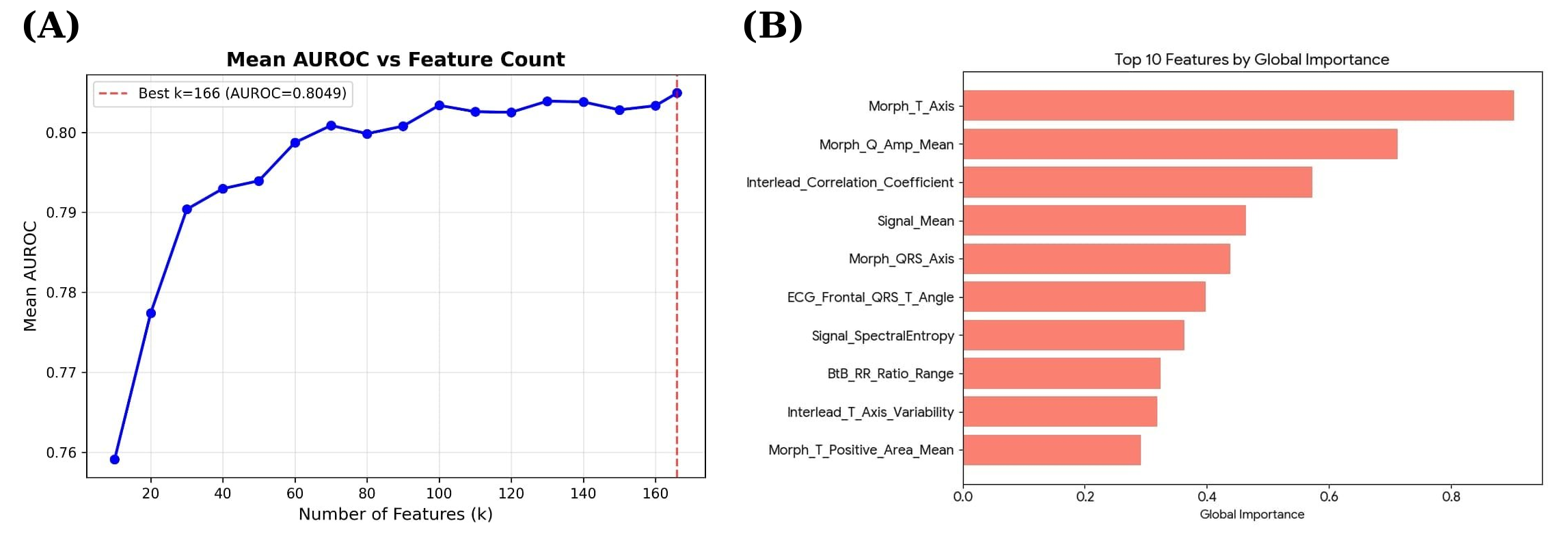}

    \caption{Engineered-feature selection and importance for Baseline~A. \textbf{(A)} Mean validation AUROC after selecting the top-\(k\) features from the global ranking. Performance improved rapidly with the first features, plateaued after roughly 60 features, and was highest when using all 166 features (AUROC \(=0.8049\)). \textbf{(B)} Top 10 features by global importance. The highest-ranked features include morphology, multi-lead, signal-statistical, and beat-to-beat descriptors.}
    \label{fig:baselineA_features}
\end{figure}

Baseline~A used a single global feature ranking across the six targets so that all one-vs-rest XGBoost models used a common feature set. The ranking combined train-set mutual information, XGBoost gain-based importance, and permutation importance with weights 0.30, 0.40, and 0.30, respectively. Figure~\ref{fig:baselineA_features} summarizes the feature-count sensitivity analysis and the highest-ranked engineered ECG features.

\section{Qualitative latent-space visualization of pretrained ECG FMs}
\label{app:tsne}

\begin{figure}[!htbp]
  \centering
  \includegraphics[width=\linewidth]{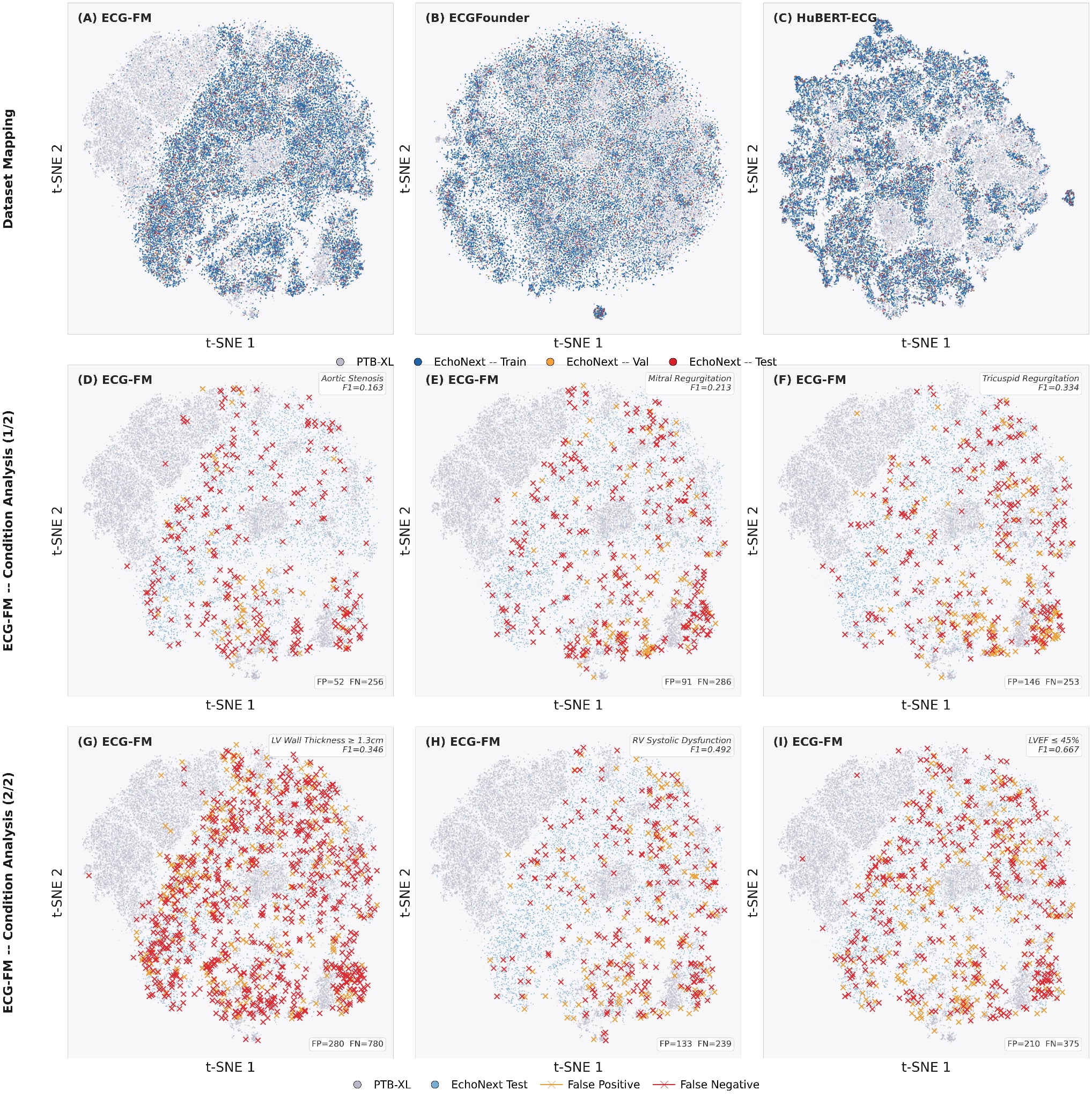}

  \caption{Qualitative t-SNE visualization of pooled embeddings from the original pretrained ECG foundation-model backbones. In Panels A--C, PTB-XL samples are shown in gray and EchoNext records with at least one modeled endpoint are colored by supervised partition (train/val/test). In Panels D--I, the pretrained ECG-FM latent space is reused to display false positives (gold) and false negatives (red) on the EchoNext test set at \(\tau=0.5\), across all six target conditions: aortic stenosis (D), mitral regurgitation (E), tricuspid regurgitation (F), LV wall thickness (G), RV systolic dysfunction (H), and reduced LVEF (I). Errors are broadly scattered with no discernible cluster boundary, consistent with the difficulty of detecting SHD from ECG waveforms alone.}
  \label{fig:appendix_tsne}
\end{figure}

Figure~\ref{fig:appendix_tsne} provides a qualitative two-dimensional visualization of pooled embeddings from the original pretrained ECG-FM, ECGFounder, and HuBERT-ECG backbones before EchoNext-specific domain adaptation or supervised fine-tuning.

\vspace{0.7em}

For each backbone, t-SNE was fitted jointly to embeddings from PTB-XL and EchoNext so that both datasets shared a common projection within the corresponding pretrained latent space \cite{wagner2020ptbxl,elias2025echonext}. Because ECG-FM yielded the strongest downstream performance in the main experiments, the lower two rows reuse its pretrained latent space to illustrate the spatial distribution of false positives and false negatives on the test set for six endpoints. These plots are intended only for qualitative inspection and should not be interpreted as evidence of class separability, calibration, or decision-boundary structure.

\section*{Code availability}
Code for model training, evaluation, and figure generation will be made publicly available at \url{https://github.com/ducdonghiem/ecgfm-shd-screening}.

\end{document}